\documentclass[runningheads]{llncs}
\usepackage{amssymb}
\usepackage[T1]{fontenc}
\usepackage{graphicx,verbatim}
\usepackage{booktabs}
\usepackage{tabularx}
\usepackage{xcolor,colortbl}
\definecolor{RowHL}{RGB}{230,242,248}

\usepackage{pifont}
\usepackage{multirow}
\usepackage{amsmath}

\usepackage{xcolor} 
\usepackage[colorlinks=true, citecolor=blue, linkcolor=blue, urlcolor=blue]{hyperref}

\usepackage{subfigure}

\begin{document}
\title{IOSVLM: A 3D Vision-Language Model for Unified Dental Diagnosis from Intraoral Scans}
\titlerunning{IOSVLM}
%

\author{Huimin Xiong\inst{1,2} \and
Zijie Meng \inst{1,2} \and
Tianxiang Hu \inst{1,2} \and 
Chenyi Zhou\inst{1} \and
Yang Feng\inst{3} \and
Zuozhu Liu \inst{1,2}
  }

\authorrunning{H. Xiong et al.}

\institute{
ZJU-UIUC Institute, Zhejiang University, Haining, 314400, China 
\email{zuozhuliu@intl.zju.edu.cn} \\ \and
Stomatology Hospital, School of Stomatology, 
Zhejiang University School of Medicine, Hangzhou, 310058, 
China \and
Angelalign Research Institute, Angel Align Inc., Shanghai, 200011, 
China 
}

  
\maketitle              
\begin{abstract}
3D intraoral scans (IOS) are increasingly adopted in routine dentistry due to abundant geometric evidence,
and unified multi-disease diagnosis 
is desirable for clinical documentation and communication. 
While recent works introduce dental vision-language models (VLMs) to enable unified diagnosis and report generation on 2D images or multi-view images rendered from IOS,
theys do not fully leverage native 3D geometry. 
Such work is necessary and also challenging, due to: (i) heterogeneous scan forms 
and the complex IOS topology, 
(ii) multi-disease co-occurrence with class imbalance and fine-grained morphological ambiguity, (iii) limited paired 3D IOS–text data.
Thus, we present IOSVLM, an end-to-end 3D VLM that represents scans as point clouds and follows a 3D encoder-projector-LLM design for unified diagnosis and generative visual question-answering (VQA), together with IOSVQA, a large-scale multi-source IOS diagnosis VQA dataset comprising 19,002 cases and 249,055 VQA pairs over 23 oral diseases and heterogeneous scan types. To address the distribution gap between color-free IOS data and color-dependent 3D pretraining, we propose a geometry-to-chromatic proxy that stabilizes fine-grained geometric perception and cross-modal alignment. A two-stage curriculum training strategy further enhances robustness. IOSVLM consistently outperforms strong baselines, achieving gains of at least +9.58\% macro accuracy and +1.46\% macro F1, indicating the effectiveness of direct 3D geometry modeling for IOS-based diagnosis.



\end{abstract}
\section{Introduction}
Intraoral scans (IOS) is rapidly becoming routine in clinical dentistry~\cite{eggmann2024recent,mangano2017intraoral}. Its high-fidelity 3D surface geometry preserves fine-grained tooth–gingiva morphology beyond conventional 2D imaging, 
enabling more consistent and auditable assessment of subtle abnormalities \cite{lian2020deep}. Clinically, a single scan often contains multiple co-existing diseases, requiring integrated multi-disease diagnosis and natural-language reporting for documentation and dentist–patient communication—beyond segmentation or single-lesion detection \cite{xiong2023tsegformer}. 

Recent dentistry work explored vision-language models (VLM) for unified diagnosis. 
DentVLM \cite{meng2025dentvlm}, DentalGPT\cite{cai2025dentalgpt}, and OralGPT \cite{zhang2025oralgpt} support visual question-answering (VQA) and generative reporting on diverse 2D dental images, providing a general interface for fusing evidence into natural-language conclusions. However, they do not directly model 3D surface geometry, where many abnormalities present as fine-grained morphological changes. OralGPT-Omni\cite{hao2025oralgpt} and ArchMap \cite{zhang2025archmap} instead render 3D scans into multi-view images and applies 2D VLMs, but this relies on view selection and can weaken spatial relationships and geometric cues.
Hence, a key gap remains: \textbf{end-to-end multi-disease diagnosis with language generation from native 3D IOS inputs.}

However, direct 3D IOS–based unified diagnosis faces challenges. First, inputs are highly heterogeneous: single-arch and occluded-arches scans acquired practically differ substantially in coverage and occlusion/contact visibility,
and IOSs exhibit complex morphology and topology, complicating representation learning~\cite{ender2019accuracy,lian2020deep,zanjani2019deep}. Second, multiple diseases often co-exist in one scan. Class imbalance and subtle geometric differences exacerbate inter-disease confusion, demanding unified reasoning from local morphology to global semantics \cite{meng2025dentvlm}. Third, 
3D IOS-text paired data and high-quality annotations is rare~\cite{TMI,rodriguez2025charnet}. 
Existing public resources are insufficient for systematic 3D VLM training and evaluation.

To this end, we propose IOSVLM, a 3D VLM model that directly consumes native 3D IOS geometry for unified 
multi-disease diagnosis and generative VQA, together with IOSVQA as a paired training and benchmark suite that covers 23 oral diseases and 2 common clinical 
settings, single-arch and occluded-arches scans, 
totalling 19,002 IOS cases and 249,055 VQA pairs. To our knowledge, it is among the most 
comprehensive IOS diagnostic VQA resources to date, explicitly capturing the real-world setting of co-existing diseases and heterogeneous inputs.

IOSVLM adopts a 3D encoder–projector–LLM framework
where the 3D encoder captures multi-scale geometric features, the projectors map them into the LLM token space, and the LLM generates communicable diagnostic outputs.
Importantly, IOS often lacks reliable color and common 3D pretraining implicitly assumes colored point clouds, naively dropping or padding color causes distribution shift and weaken cross-modal alignment. We thus introduce the geometry-to-chromatic proxy to compensate for that,
to improve fine-grained morphology encoding and language alignment. IOSVLM is trained with a curriculum-style schedule: pre-training on larger but noisy supervision to build 3D perception and geometry–language alignment, and fine-tuning on higher-quality
supervision to enhance reliability and interpretability under realistic label noise and limited explanations. 
Experiments show that IOSVLM substantially outperforms strong state-of-the-art baselines, improving macro accuracy by at least 9.58\% and macro F1 by at least 1.46\%. 
Our main contributions are:
\begin{enumerate}
    \item We construct the first large-scale multi-source IOS diagnostic VQA dataset that supports multiple scan type inputs for multi-disease diagnosis.
    \item We introduce the first end-to-end VLM that takes native 3D IOS geometry as input for unified diagnosis, achieving clear performance advantages.
    \item Geometry-to-chromatic proxy is proposed to bridge the distribution gap between color-free IOSs and color-dependent point-cloud pretraining, improving fine-grained geometry encoding and cross-modal alignment.
\end{enumerate}

\begin{table}[t]
  \centering
  \caption{Statistics of IOSVQA and its source datasets. S: single-arch. O: occluded-arch. SD: single-arch disease. OD: occluded-arch disease. \# denotes the case number.}
  \label{tab:ios_dataset_summary}
  \small
  \setlength{\tabcolsep}{3.5pt}
  \renewcommand{\arraystretch}{1.15}
  \begin{tabularx}{\columnwidth}{@{} l c l c c c c c @{}}
    \toprule
    \textbf{Dataset} & \textbf{Privacy} & \textbf{Source} & \textbf{IOS types} &
    \textbf{\#SD} & \textbf{\#OD} &
    \textbf{\#cases} & \textbf{\#VQA} \\
    \midrule
    MaloccIOS       & Private & China & S+O & 2    & 11  & 14,630 & 208,370 \\
    DiseaseIOS      & Private & China & S   & 8    & --- & 4,172  & 33,376 \\
    Bits2Bites      & Public  & Italy & O   & ---  & 5   & 200    & 7,309 \\
    \midrule
    IOSVQA          & ---     & Mixed & S+O & 8    & 15  & 19,002  & 249,055 \\
    \bottomrule
  \end{tabularx}
\end{table}

\section{Method}
\subsection{Problem Formulation}
IOSVLM targets two common IOS inputs: single-arch scans 
and occluded-arches scans (refer to Fig. \ref{case_study}).
Let $\mathcal{D} = \{1, \dots, C\}$ denote the set of oral diseases, where $C = 23$.
Each disease $d \in \mathcal{D}$ corresponds to a multi-class label set $Y_d$. Given an input pair \(\bigl(\widetilde{M}, q_d\bigr)\) with IOS mesh $\widetilde{M}$ and question \(q_d\), the diagnosis is formulated as a generative VQA task that produces an answer \(A_d \in \{y,\,(y,r)\}\) ($y \in Y_d$), i.e., a label \(y\) or a label--rationale pair \((y,r)\). Since multiple diseases may co-exist within a single scan, each scan is paired with multiple disease questions. Training and evaluation are conducted over scan--disease pairs as disease-level instances.


\begin{figure}[tbp]
\subfigure[Occluded-arches]{
\label{dist_occu}
\includegraphics[width=0.46\linewidth]{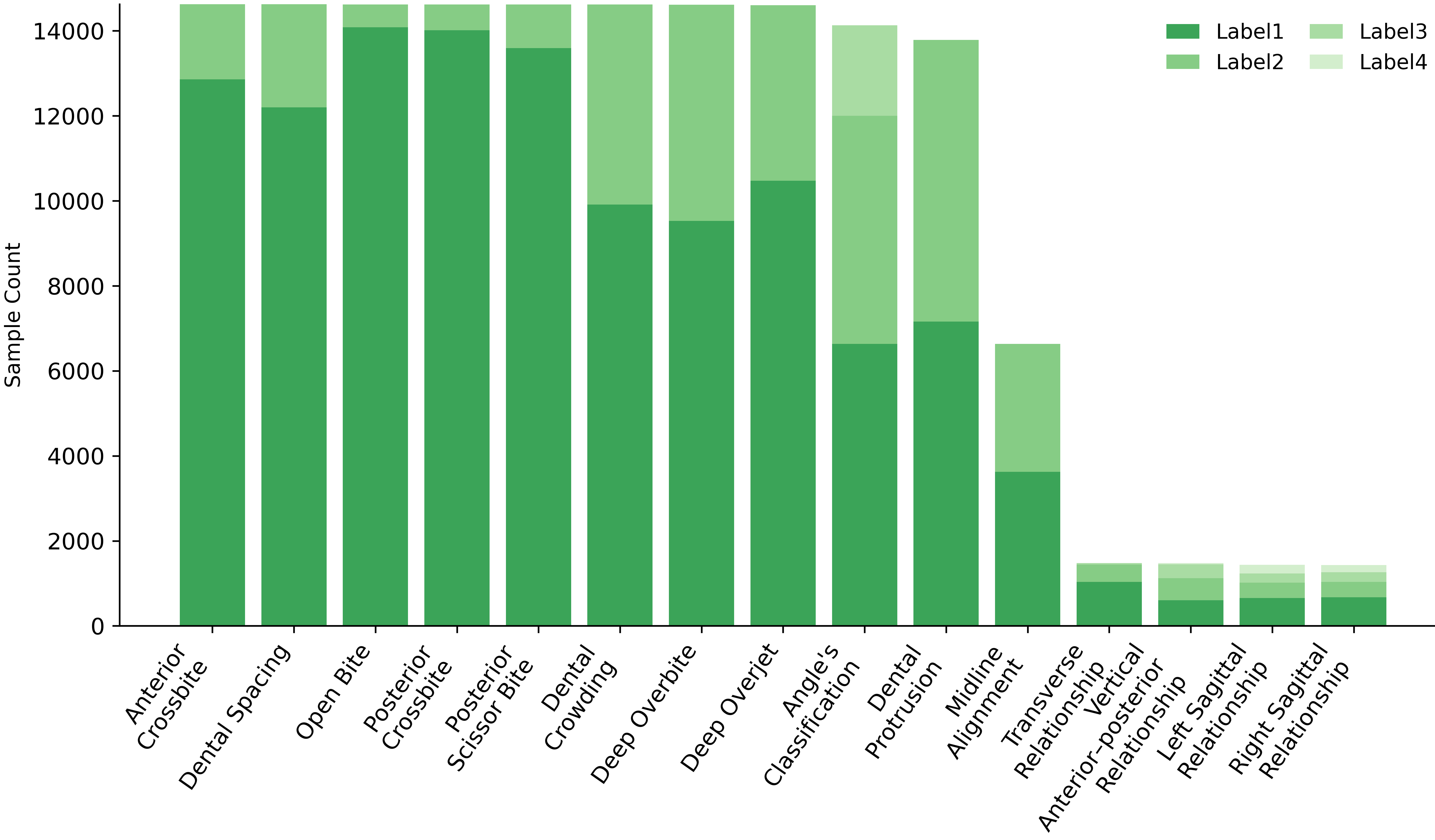}
}
\subfigure[Single-arch]{
\label{dist_single}
\includegraphics[width=0.25\linewidth]{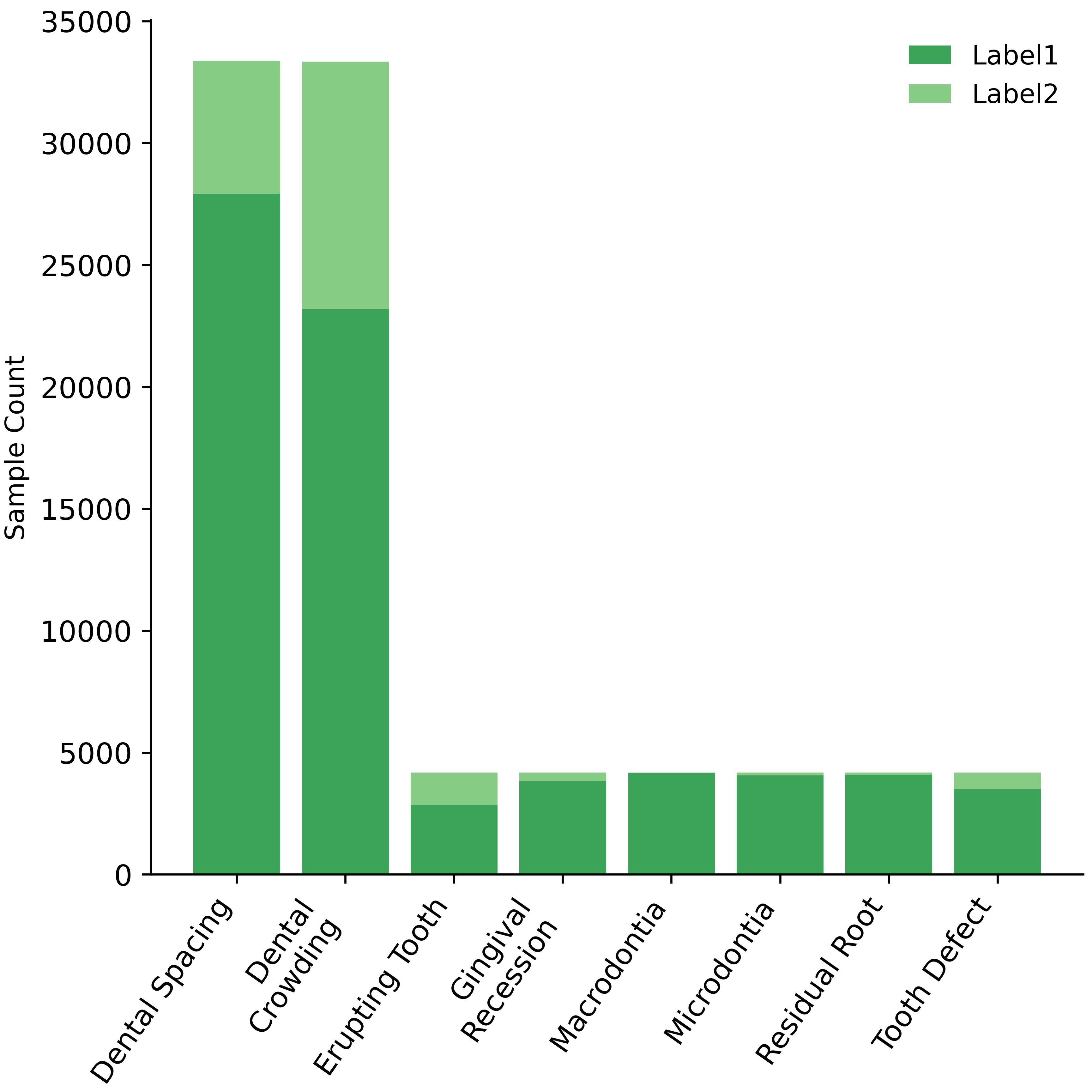}
}
\subfigure[GCP]{
\label{normal_color}
\includegraphics[width=0.21\linewidth]{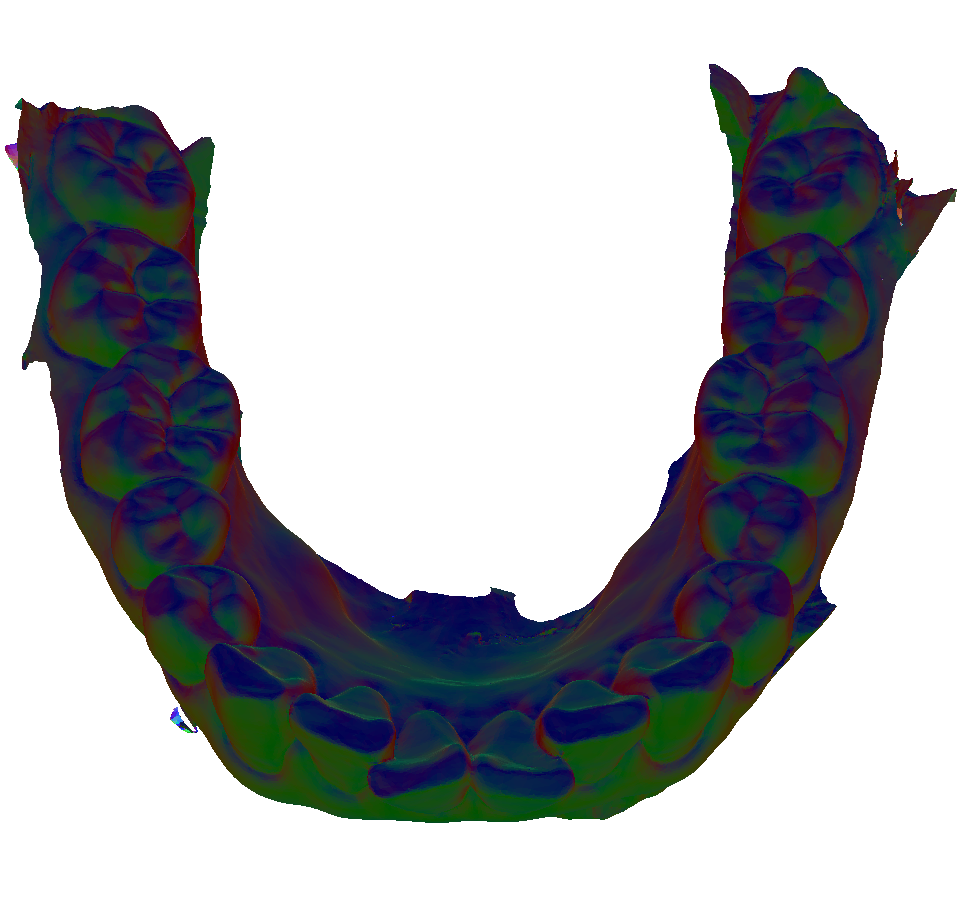}
}
\caption{Disease statistics of occluded-arches IOSs (a) and single-arch (b) on IOSVQA dataset, and visualization of normal-based geometry-to-chromatic proxy (GCP) (c).}
\label{disease dist & normal color}
\end{figure}

\subsection{IOSVQA Dataset Construction}
We first construct a large-scale VQA dataset IOSVQA.
It comprises 19,002 IOS cases and 249,055 QA pairs spanning 23 disease categories, with disease statistics detailed in Fig. ~\ref{disease dist & normal color}. Data were aggregated from three sources: MaloccIOS, DiseaseIOS, and Bits2Bites \cite{borghi2025bits2bites} (Table ~\ref{tab:ios_dataset_summary}). To make geometric representations comparable across sources, IOSs were globally registered to standardize orientation and relative maxillomandibular pose. Each case provides one IOS paired with a disease label to form QA samples; questions were sampled randomly from a predefined list and answers correspond predominantly to disease labels.

\textbf{Disease Label Consolidation.} 
For MaloccIOS, 2D imaging diagnoses were extracted from clinical reports. A randomly selected subset of 557 cases was manually corrected by 28 orthodontists, yielding 7,628 high-quality samples, while the remaining 200,742 samples contain partial label noise. Senior dentists defined rule-based mappings to convert 2D diagnoses into IOS labels. Labels for DiseaseIOS and Bits2Bites were annotated by 5 and 1 orthodontic experts, respectively, and are treated as high quality. 





\textbf{Data Split and Rationale Strategy.} IOSVQA is divided into Stage-1 training, Stage-2 training, and testing, with each stage covering all three data sources to maintain disease diversity. The 7,628 high-quality MaloccIOS samples are fully assigned to Stage-2.
For interpretability, approximately 50\% of Stage-2 samples are augmented with GPT-4o–generated chain-of-thought (CoT) rationales. Motivated by prior evidence that limited CoT supervision can induce reasoning behavior~\cite{zelikman2024star}, we apply rationale annotation to a subset of high-quality data to balance cost, efficiency, and reasoning performance. To mitigate scale imbalance, the smaller Bits2Bites dataset is augmented by pairing each case with 9 and 3 questions for Stage-1 and Stage-2 construction, respectively.

\begin{figure}[!t]
\includegraphics[width=\textwidth]{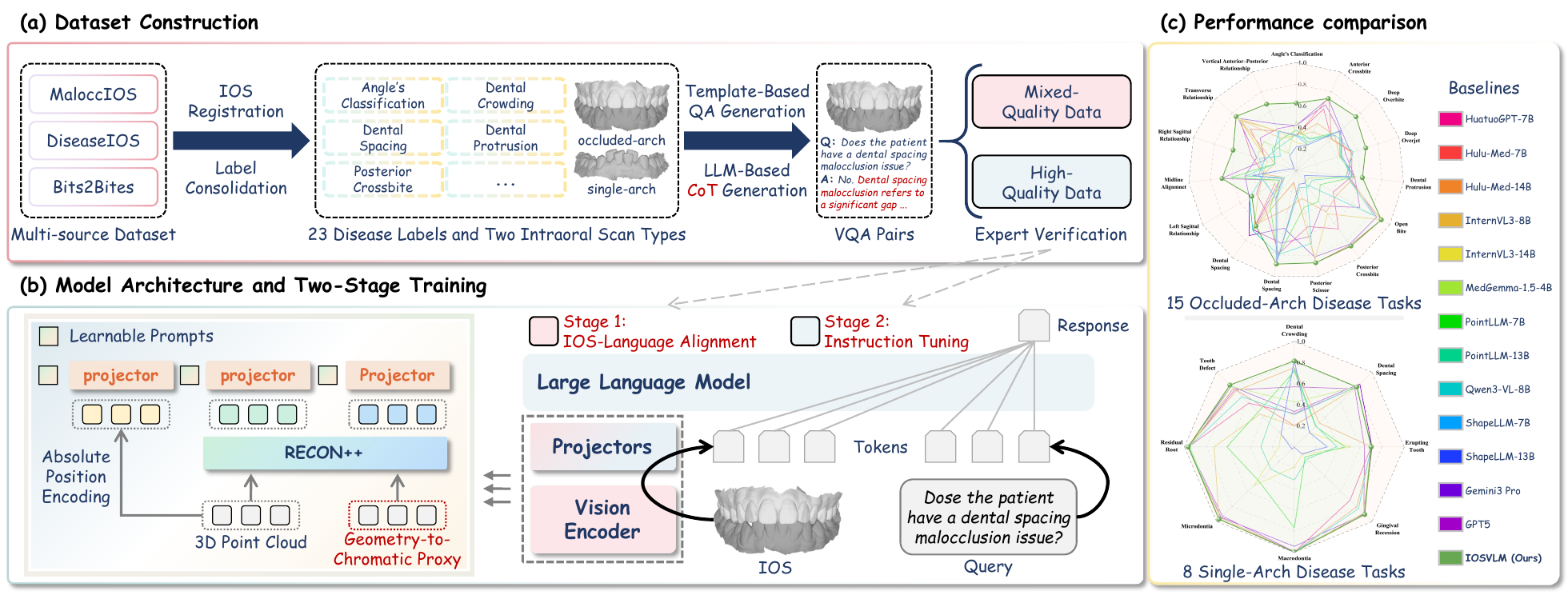}
\caption{\textbf{Overview of our framework and dataset construction.} }
\label{pipeline}
\end{figure}

\subsection{Model Architecture}
We adopt the multi-modal VLM architecture that combines a large-scale pretrained 3D encoder with multi-branch projectors and LLM (Fig. \ref{pipeline}). The 3D encoder injects strong geometric priors, while the structured projectors map multi-granularity visual features into semantic tokens aligned with the LLM. The LLM then performs reasoning to produce diagnostic outputs, enabling unified perception and explainable diagnosis across heterogeneous IOS inputs.

To map the raw IOS mesh $\widetilde{M}$ into the semantic space of LLM, we first convert it to a point cloud $\widetilde{P}$ by taking the gravity center point of each face. $\widetilde{P}$ is then randomly down-sampled to obtain $P$ with $N$ points. 
A pre-trained 3D point-cloud encoder ReCon++ \cite{qi2024shapellm} extracts complementary 
absolute position embeddings \(F_{\mathrm{ape}}\), local geometric features \(F_{\mathrm{local}}\), and a global descriptor \(F_{\mathrm{global}}\). 
They are mapped into the LLM token space via three dedicated MLP projectors \(\phi_{\mathrm{ape}}, \phi_{\mathrm{local}}, \phi_{\mathrm{global}}\), and each projected feature 
is concatenated with a corresponding learnable visual prompt \(V_{\mathrm{ape}}, V_{\mathrm{local}}, V_{\mathrm{global}}\) to balance their contributions:
\begin{align}
F_p &= \big[ V_{\mathrm{ape}} ; \phi_{\mathrm{ape}}\big(F_{\mathrm{ape}}\big) ; V_{\mathrm{local}} ; \phi_{\mathrm{local}}\big(F_{\mathrm{local}}\big); V_{\mathrm{global}}; \phi_{\mathrm{global}}\big(F_{\mathrm{global}}\big)].
\end{align}
The fused point tokens $F_p$ are concatenated with text tokens and jointly fed into the LLM, enabling unified language–visual representations.

\textbf{Geometry-to-Chromatic Proxy.}
Point-cloud related models are often pretrained with \(\mathrm{xyz}+\mathrm{RGB}\) inputs.
In practical IOS pipelines, downstream storage and exchange typically preserve geometry while discarding or de-emphasizing color/texture, making appearance cues unavailable. Removing RGB thus introduces a pretraining mismatch and discards channels the encoder has learned to exploit. Moreover, RGB is often beneficial not due to semantic color, but as a local separability cue that supports boundary discovery
and stable local matching. We therefore propose \emph{Geometry-to-Chromatic Proxy (GCP)}: a geometry-derived proxy that mimics this separability component, enabling better reuse of color-pretraining priors while remaining purely geometry-driven.

We instantiate GCP using surface normals, which encode local surface orientation and capture curvature changes etc.
For each point \(i\) with normal \(n_i\), 
we define a flip-robust mapping
$\mathrm{GCP}(n_i)=\left|\frac{n_i}{\|n_i\|_2}\right|\in\mathbb{R}^3$.
The \(\ell_2\) normalization standardizes magnitude, and the absolute value removes sign ambiguity, avoiding spurious color inversions. Mapping \(\mathrm{GCP}(n_i)\) onto the IOS as pseudo-colors (Fig.~\ref{normal_color}) yields spatially coherent patterns with clear transitions at anatomical boundaries, supporting that GCP provides structured, locally discriminative cues analogous to RGB. 
This formulation is extensible: other geometry descriptors (e.g., curvature) may serve as alternative proxies and yield 
gains.
Overall, GCP provides a geometry-driven substitute for the discriminative component of RGB, improving fine-grained geometric perception 
when color is 
unreliable.

\subsection{Training Strategy}

We adopt a curriculum-style two-stage training: Stage-1 learns robust 3D geometry and geometry-language alignment, while Stage-2 improves diagnosis and generation under higher-quality (partially rationale-augmented) supervision.

\textbf{Stage-1.} We train the 3D encoder and projectors and freeze the LLM, leveraging the large-scale Stage-1 data to build strong geometric representations and stable token alignment. Point features with GCP are used to reduce the gap to color-dependent point-cloud pretraining.

\textbf{Stage-2.} We freeze the 3D encoder and fine-tune the projectors and LLM with LoRA, leveraging higher-quality annotations to further enhance semantic modeling and generative reasoning, while preventing degradation of the learned geometric representations. A unified generative objective is applied: label-only samples supervise \(y\), while rationale samples supervise \((y,r)\).

\begin{table}[t]
\centering
\caption{Comparison with baseline methods on multi-disease IOS diagnosis. The best result is bolded, while the second best result is underlined.
}
\label{tab:baselines}
\small
\setlength{\tabcolsep}{7pt}
\renewcommand{\arraystretch}{1.15}

\resizebox{0.8\textwidth}{!}{\begin{tabularx}{\linewidth}{@{} l c *{5}{>{\centering\arraybackslash}X} @{}}
\toprule
\textbf{Models} & \textbf{Input} &
\textbf{Acc} & \textbf{F1} & \textbf{Preci} & \textbf{Recall}  & \textbf{PR}\\
\midrule

GPT-5 ~\cite{singh2025openai}        & Multi-View Images & 62.26 & 44.73 & 49.41  & 49.20  & 99.84  \\
Gemini 3 Pro ~\cite{gemini3pro_modelcard_2025} & Multi-View Images & \underline{67.65} & \underline{48.93} & \textbf{56.78} & \underline{51.85}  & 99.71 \\
\midrule[\heavyrulewidth]

Qwen3VL-8B \cite{bai2025qwen3}  & Multi-View Images & 57.88 & 36.46 & 45.21 & 46.58 & 100 \\
InternVL3.5-8B \cite{wang2025internvl3} & Multi-View Images & 56.23 & 34.77 & 40.36 & 45.47 & 100 \\
InternVL3.5-14B \cite{wang2025internvl3} & Multi-View Images & 35.68 & 24.61 & 29.14 & 46.87 & 100 \\
\midrule[\heavyrulewidth]

MedGemma-1.5 \cite{sellergren2025medgemma} & Multi-View Images & 52.21 & 31.31 & 36.58 & 45.25 & 100 \\
HuatuoGPT-V-7B \cite{chen2024towards} & Multi-View Images & 59.26  & 39.14  & 40.33  & 45.56 & 100  \\
HuluMed-7B \cite{jiang2025hulu} & Multi-View Images & 61.21 & 34.68  & 40.09 & 45.12 & 100  \\
HuluMed-14B \cite{jiang2025hulu} & Multi-View Images & 53.72 & 36.87 & 41.63 & 46.58 & 100  \\
\midrule[\heavyrulewidth]

PointLLM-7B \cite{xu2024pointllm} & 3D Point Cloud & 43.03 & 34.18 & 44.62  & 44.80 & 99.85   \\
PointLLM-13B \cite{xu2024pointllm} & 3D Point Cloud & 34.57 & 26.76 & 42.04 & 45.03 & 96.78   \\
ShapeLLM-7B \cite{qi2024shapellm} & 3D Point Cloud & 30.27 & 19.23  & 14.66 & 44.95  & 100 \\
ShapeLLM-13B \cite{qi2024shapellm} & 3D Point Cloud & 26.13 & 17.78 & 13.61 & 45.10  & 100 \\
\rowcolor{RowHL}
\textbf{IOSVLM(Ours)} & 3D Point Cloud & \textbf{77.23} & \textbf{50.39} & \underline{52.19} & \textbf{52.96} & \textbf{100} \\
\bottomrule
\end{tabularx}}
\end{table}

\section{Experimental Results}

\subsection{Experimental Setup}
We evaluate on IOSVQA, with 229,943/15,598/5,884 samples for Stage-1 training, Stage-2 training, and testing, respectively. 
We report Macro Accuracy (Acc), Macro F1 (F1), Precision (Preci), and Recall, averaged over tasks, and additionally Parsing Rate (PR), measuring the fraction of generated answers that can be parsed into a valid label.
IOSVLM uses LLM from Qwen3VL-8B-Instruct \cite{bai2025qwen3} with N=10,000 points and 32 learnable visual prompts. 

\subsection{Comparison with State-of-the-Arts}
Table~\ref{tab:baselines} compares IOSVLM with 4 categories of representative baselines:
(i) proprietary multimodal LLMs ~\cite{singh2025openai,gemini3pro_modelcard_2025}, (ii) open-source general-purpose 2D MLLMs ~\cite{bai2025qwen3,wang2025internvl3}, (iii) open-source medical 2D MLLMs ~\cite{sellergren2025medgemma,chen2024towards,jiang2025hulu}, and (iv) open-source general 3D MLLMs operating on point clouds ~\cite{xu2024pointllm,qi2024shapellm}. For fair comparison, all 3D baselines use the same point-cloud preprocessing as IOSVLM, 2D baselines receive conventional multi-view renderings of IOS: 5 standard intraoral views for occluded-arch scans 
and 4 views for single-arch scans. 





\textbf{Overall performance.}
IOSVLM achieves the best overall results, obtaining the highest Acc (77.23\%) and F1 (50.39\%), as well as the best Recall (52.96\%). Notably, IOSVLM outperforms all open-source 2D MLLMs by a large margin (at least +16.02\% Acc/+11.25\% F1) and substantially surpasses open-source 3D MLLMs (at least +34.20\% Acc/+16.21\% F1), highlighting the advantage of directly modeling native 3D IOS geometry.

\textbf{Comparison to proprietary MLLMs.}
Compared with GPT-5, IOSVLM improves by +14.97\% Acc and +5.66\% F1; compared with Gemini 3 Pro, it yields +9.58\% Acc and +1.46\% F1 and achieves higher recall (52.96\% vs.\ 51.85\%). Notably, IOSVLM uses only an 8B-scale LLM, yet surpasses these substantially larger proprietary MLLMs, highlighting that our gains primarily come from geometry-aware representation and alignment rather than model size. Gemini 3 Pro attains the best precision (56.78\%), while IOSVLM provides a more balanced profile with the strongest accuracy-F1-recall, which is crucial for multi-disease diagnosis under class imbalance and fine-grained ambiguity.

\textbf{Parsing rate (PR).}
IOSVLM reaches 100\% macro PR, indicating consistently parsable outputs in the task-specific label space, while proprietary models show slightly lower PR (99.84\%/99.71\%). We observe that, despite near-perfect PR, proprietary models still exhibit sporadic empty outputs that break parsability. Together, these results demonstrate that IOSVLM not only improves diagnostic performance but also produces more reliably structured predictions and informative output for large-scale multi-task evaluation.

\begin{table}[t]
\centering
\caption{Ablation studies. V/P/L denote 3D vision encoder, projector, and LLM, respectively. "\(*\)" indicates tuning with rationales. GCP: Geometry-to-Chromatic Proxy.}
\label{tab:ablations}
\small
\setlength{\tabcolsep}{7pt}
\renewcommand{\arraystretch}{1.15}

\resizebox{0.8\textwidth}{!}{\begin{tabularx}{\linewidth}{@{} c c *{5}{>{\centering\arraybackslash}X} @{}}
\toprule
\textbf{Ablation part} & \textbf{Method choice} &
\textbf{Acc} & \textbf{F1} & \textbf{Preci} & \textbf{Recall} & \textbf{PR} \\
\midrule

\multirow{3}{*}{\textbf{LLM}} & Qwen3-4B \cite{yang2025qwen3}   & 71.10  & 39.22 & 35.55 & 45.15 & 100 \\
\multicolumn{1}{c}{}         & Qwen3-8B \cite{yang2025qwen3}    & 70.73 & 38.84 & 36.42 & 45.26 & 100 \\
\multicolumn{1}{c}{}         & \textbf{Qwen3VL-8B}  & 67.02 & \textbf{43.10} & \textbf{48.94} & \textbf{46.70} & 95.08 \\
\midrule[\heavyrulewidth]

\multirow{2}{*}{\textbf{GCP}} & \ding{55} & 67.02 & 43.10 & 48.94 & 46.70 & 95.08  \\
\multicolumn{1}{c}{}            & \ding{51} & \textbf{72.28}  & \textbf{48.06} & \textbf{51.25} & \textbf{49.82} & \textbf{97.00} \\
\midrule[\heavyrulewidth]

\multirow{5}{*}{\centering\parbox{2.8cm}{\centering\textbf{Training\\mode}}} & Stage 1: V \& P  & 72.28  & 48.06 & 51.25 & 49.82 & 97.00  \\
\multicolumn{1}{c}{}                    & *Stage 2: P \& L   & \textbf{77.23} & \textbf{50.39} & \textbf{52.19} & \textbf{52.96} & \textbf{100} \\
\multicolumn{1}{c}{}                    & Stage 2: P \& L   & 77.92 & 50.35 & 51.81 & 53.61 & 99.74 \\
\cmidrule(lr){2-7}  
\multicolumn{1}{c}{}                   & Stage 1: V \& P \& L & 75.38 & 47.79 & 49.73 & 51.07 & 99.97  \\
\multicolumn{1}{c}{}                   & *Stage 2: P \& L   & 75.72  & 50.78 & 52.37 & 52.62 & 100 \\
\bottomrule
\end{tabularx}}
\end{table}

\begin{figure}[!t]
\includegraphics[width=\textwidth]{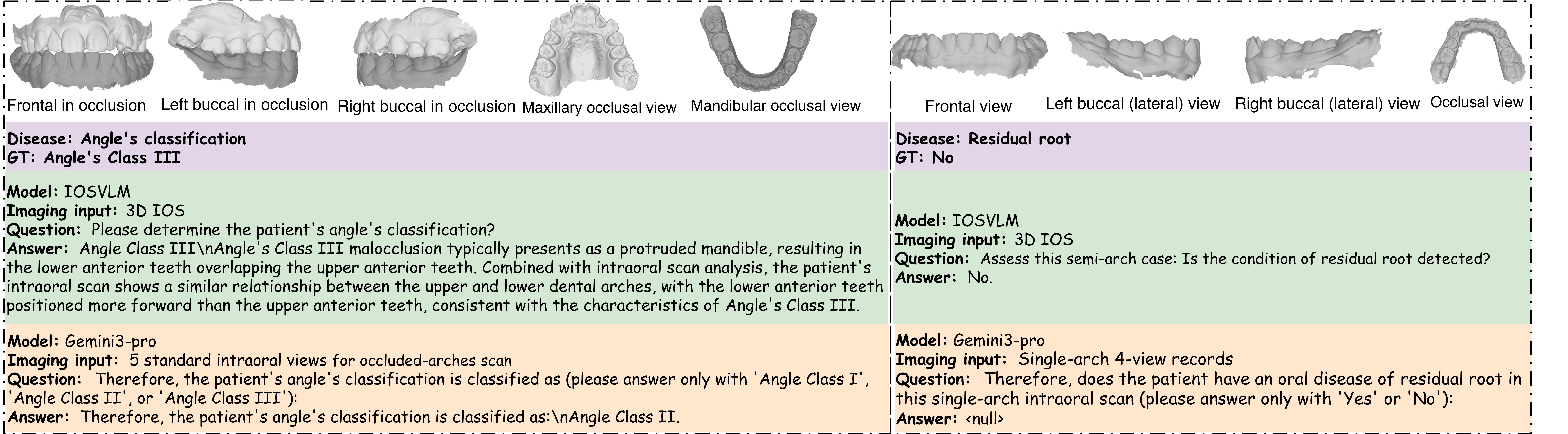}
\caption{Qualitative examples of generated diagnostic responses. 
} 
\label{case_study}
\end{figure}

\subsection{Ablation Studies}

\textbf{Effect of Geometry-to-Chromatic Proxy (GCP).}
In Table \ref{tab:ablations}, ``\(\times\)'' denotes using a constant ``white'' color (i.e., no separability cue), while ``\(\checkmark\)'' replaces the color channels with the normal-based proxy. 
GCP yields consistent gains (\(+5.26\%\) Acc and \(+4.96\%\) F1), supporting our hypothesis that the discriminative component of RGB, not its semantics, is what benefits pretrained encoders. This also validates normal vectors as an effective instantiation of GCP, and suggests the proxy can be further extended to other geometric cues (e.g., curvature).

\textbf{Training strategy.}
Our two-stage curriculum achieves strong performance, with the default setting yielding the best overall accuracy and F1. Interestingly, training V\&P\&L in Stage-1 outperforms V\&P initially, but becomes slightly inferior after Stage-2 in the Acc–F1 trade-off. We attribute this to early LLM updates under mixed-quality supervision, which may bias generation and limit subsequent high-quality adaptation. Nevertheless, all variants are competitive, suggesting that native 3D IOS input with a VLM pipeline forms a strong foundation, while the staged curriculum provides more stable optimization for multi-disease ambiguity.

\textbf{Rationale supervision: reliability over raw F1.}
Using rationales in Stage-2 yields comparable macro-F1 to label-only tuning, suggesting that rationales mainly regularize generation rather than improving geometric discriminability. Importantly, rationale tuning improves output reliability/usability: it maintains diagnostic performance while achieving higher macro PR, 
, and we qualitatively observe fewer degenerate behaviors such as repeated labels. This indicates improved controllability and better-calibrated free-form answers, which is crucial for downstream clinical pipelines.

\textbf{Choice of LLM.}
Among LLM candidates, Qwen3VL-8B yields the best macro-F1 under our setting. Although its macro accuracy is slightly lower than the other LLMs, we observe that in our highly imbalanced tasks those LLMs tend to collapse to majority-class predictions, inflating accuracy but hurting minority coverage. In contrast, Qwen3VL-8B exhibits better class-coverage awareness and less majority bias, leading to a more balanced precision-recall profile.

\subsection{Qualitative Analysis}
In Fig. \ref{case_study}, IOSVLM shows
more accurate diagnoses and more reliable, parsable outputs than the strong Gemini 3 Pro,
supporting that directly modeling native 3D IOS geometry improves sensitivity to fine-grained morphological cues and enhances output robustness for clinical use.



\section{Conclusion}
We presented IOSVLM, an end-to-end 3D VLM for unified multi-disease IOS diagnosis and generative VQA, together with IOSVQA, a large-scale benchmark spanning 23 oral diseases and heterogeneous scan types. We introduce a geometry-to-chromatic proxy and a curriculum-style training strategy to better leverage color-dependent point-cloud pretraining and improve robustness. Experiments demonstrate clear gains over strong baselines, supporting direct 3D geometry modeling for practical IOS-based clinical use.

%
%
%
%
\bibliographystyle{splncs04}
\bibliography{ref}

\end{document}